\documentclass[conference]{IEEEtran}
\IEEEoverridecommandlockouts
\usepackage{cite}
\usepackage{amsmath,amssymb,amsfonts}
\usepackage{algorithmic}
\usepackage{graphicx}
\usepackage{textcomp}
\usepackage{xcolor}
\usepackage{graphicx}
\usepackage{subcaption}
\usepackage{hyperref}
\usepackage{multirow} 
\usepackage{diagbox}
\usepackage{booktabs}
\usepackage{marvosym}
\usepackage[numbers,sort&compress]{natbib}
\def\BibTeX{{\rm B\kern-.05em{\sc i\kern-.025em b}\kern-.08em
    T\kern-.1667em\lower.7ex\hbox{E}\kern-.125emX}}
\begin{document}

\title{LingLan: An Advancing Traditional Chinese Medicine Diagnosis LLM with Multimodal Data}

% \author{Anonymous ICME submission}
\author{\textbf{Zheng Chen}$^1$, \textbf{Zhicheng Du}$^1$, \textbf{Haoxuan Li}$^1$, \textbf{Yingshan Liang}$^1$, \textbf{Peiwu Qin}$^{2\dag}$\\

$^1$Tsinghua Shenzhen International Graduate School, Tsinghua University, Shenzhen, China \\

$^2$Guangdong Provincial Laboratory of Traditional Chinese Medicine Hengqin, Guangdong, China

\thanks{$^{2\dag}$Corresponding author: pwqin1979@gmail.com.}}

\maketitle

\begin{abstract}
Though artificial intelligence (AI) increasingly transforms modern medicine, its integration into Traditional Chinese Medicine (TCM) has been relatively slow, primarily due to TCM's reliance on holistic, subjective diagnostic methods—namely Inspection, Auscultation and Olfaction, Inquiry, and Palpation(I-AOI-P)—which are difficult to align with quantitative, standardized medical systems. In this work, we introduce a Unification Framework for Multimodal Data (UFMD), which automatically processes tongue and pulse images into structured, clinically standard descriptions, integrating multi-source diagnostic information into a unified digital record of I-AOI-P process. Building on this structured data, we create LingLan-14B, a TCM-specific large language model fine-tuned via supervised learning to emulate the diagnostic logic and workflow of I-AOI-P process. Experimental results show that our method significantly enhances diagnostic accuracy, achieving a relative improvement of 103.5\% over the baseline (62.72\% vs. 30.82\%) and reaching an F1-score of up to 82\%.
\end{abstract}

\begin{IEEEkeywords}
Multimodal Large Language Model, Traditional Chinese Medicine Diagnosis, Supervised Fine-Tuning
\end{IEEEkeywords}

\section{Introduction}
\label{sec:intro}

\begin{figure}
    \centering
    \includegraphics[width=\linewidth]{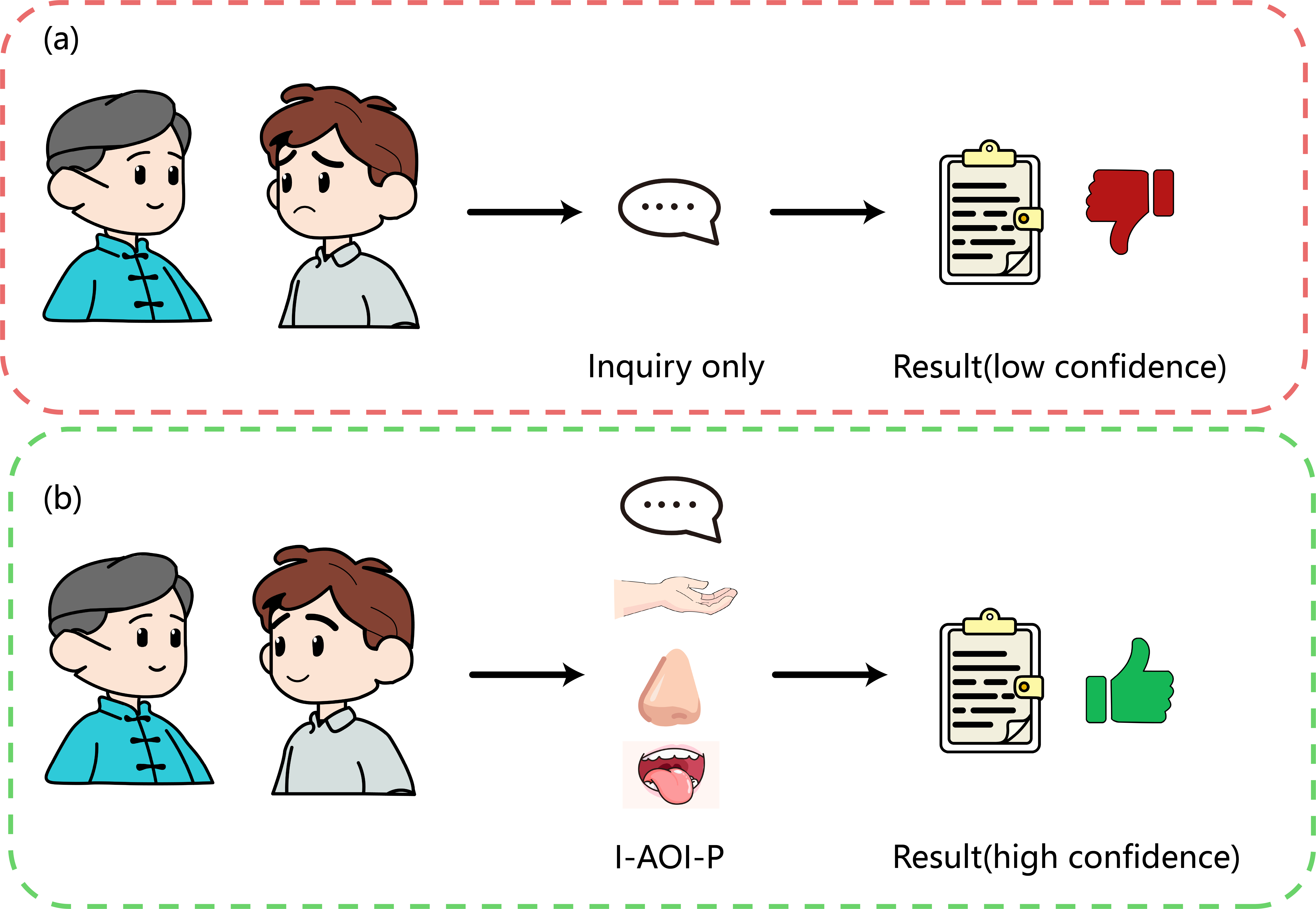}
    \caption{(a) Diagnosis with only inquiry will result in low-reliability results. (b) Diagnosis with I-AOI-P Diagnosis will yield more accurate results. }
    \label{fig:diagnosis}
\end{figure}

In recent years, the rapid advancement of artificial intelligence (AI) has profoundly reshaped the diagnostic and treatment models, as well as the service ecosystem, of modern medicine. From precision surgical robots to large language model (LLM) based auxiliary diagnostic systems, AI has progressively integrated into the entire chain of healthcare services~\cite{liu2024surveymedicallargelanguage,wang2024directdiagnosisllmbasedmultispecialist,wang2023coadautomaticdiagnosissymptom,bao2023discmedllmbridginggenerallarge,wei-etal-2018-task}. However, Traditional Chinese Medicine (TCM), with its emphasis on holism, syndrome differentiation and treatment, and its reliance on the subjective experiential diagnostics of \textit{Inspection, Auscultation and Olfaction, Inquiry, and Palpation} (I-AOI-P, shown in Fig.~\ref{fig:diagnosis}) struggles to fully synchronize with the modern medical system characterized by quantitative analysis and standardized pathways, resulting in a relatively lagging pace of development.

Concurrently, against the backdrop of an accelerating aging population and the increasing burden of chronic diseases, societal demand for TCM's "preventive treatment of disease" philosophy and personalized health management is steadily growing, which highlights the urgency of innovating TCM diagnostic and treatment models~\cite{fmicb}. However, current AI-assisted TCM diagnosis faces challenges of insufficient accuracy and stability, making it difficult to support reliable clinical application. On one hand, TCM syndrome differentiation depends on the comprehensive judgment of multi-modal information such as tongue and pulse images, yet existing algorithms focus more on inquiry information, encountering bottlenecks in fusing multi-source heterogeneous data and modeling nonlinear relationships. On the other hand, publicly available TCM datasets are limited in scale and uneven in quality, failing to meet the demand for high-granularity, multi-dimensional data required for individualized diagnosis and treatment. Therefore, there is a pressing need to bridge this gap by combining algorithmic innovations with the construction of high-quality data resources.

To address the aforementioned challenges, we propose an innovative multi-modal large language model (MLLM) based framework for intelligent diagnosis in TCM. The core innovation of this framework lies in establishing an automated, structured pipeline that processes multi-source information into a unified knowledge representation. Specifically, the framework acquires the patient's tongue diagnosis images and pulse diagnosis waveform images, which are directly input into the system. It then outputs standardized textual descriptions of tongue manifestations (e.g., pale-red tongue proper, thin white coating) and pulse manifestations (e.g., wiry pulse, thready pulse) that conform to TCM clinical terminology, thereby objectifying and digitizing the information obtained from inspection and palpation. Ultimately, data from the I-AOI-P method are integrated into a unified, structured case representation, forming a complete digital record of the I-AOI-P process.

Based on the digital records, we perform post-training on a pre-trained LLM, which involves domain-adaptive fine-tuning aimed at aligning the model's internal reasoning mechanisms with the specialized knowledge and diagnostic logic of TCM. By learning from the unified structured representations, the model is enhanced to perform better in TCM analysis.
The main highlights of our contributions are as follows:
\begin{itemize}
  \item \textbf{Unification Framework for Multimodal Data (UFMD)}: We present a novel framework for unifying TCM multimodal clinical record data, which leverages a MLLM and incorporates human-in-the-loop verification to integrate TCM medical record information from diverse sources into structured data. 
  \item \textbf{LingLan-14B: TCM Diagnosis LLM}: Based on the unique I-AOI-P diagnostic process of TCM, we reframe the supervised fine-tuning (SFT) of the model from the perspective of its data framework, enabling the LLM to more effectively learn the underlying logic and workflow of I-AOI-P process. We will release the weight parameters of LingLan-14B after acceptance.
  
%   \href{https://modelscope.com}{[LingLan-14B]}.
  \item \textbf{Experiment Study}: Performance and ablation experiments demonstrate that SFT based on I-AOI-P diagnosis greatly enhances the model's ability to diagnose TCM, achieving a high improvement of 103.5\% over the baseline and achieving an F1-score as high as 82\%.
\end{itemize}

\section{RELATED WORK}

\textit{1) Multimodal Data Unification}: Current approaches to TCM data acquisition frequently integrate database compilation with manual curation. For instance, Liu et al.~\cite{TCM-KDIF} aggregate prescription, herbal, ingredient, and target data from 14 authoritative TCM databases. They employ natural language processing (NLP), specifically the BioBERT model, to extract and validate entity relationships from over 30 million documents, supplemented by manual curation of clinical records to ensure comprehensiveness. Similarly, Li et al.~\cite{LTM-TCM} integrate multiple authoritative databases, manually curate clinical and ancient text data, and apply biomedical NLP techniques to extract and refine relationships from extensive literature. However, the inefficiency in cleansing large-scale real-world data and the inherently time-intensive nature of manual curation processes still exist.

 \textit{2) TCM Diagnosis Improvement}: Many studies urge to improve TCM diagnosis with AI~\cite{icic,isai,ruan}. He et al.~\cite{OpenTCM} propose the OpenTCM system, which improves the accuracy and reliability of TCM diagnosis by combining LLM, knowledge graphs (KG), and graph-based retrieval enhancement generation (GraphRAG). Yang et al.~\cite{TCM-GPT} construct a large-scale TCM corpus and develop the TCM-GPT-7B model, which significantly improve its accuracy in TCM examinations and diagnostic tasks. However, these approaches have overlooked the comprehensive consideration of I-AOI-P diagnosis, which results in limited generalization capability of the models in practical applications.

\section{Methodology}
\label{Metho}
\subsection{Unification Framework for Multimodal Data}

With the deep integration of information technology into the medical field, the format of TCM clinical records is undergoing a fundamental shift from traditional paper-based documentation to comprehensive electronic systems. However, a prominent challenge in this transition is the significant heterogeneity of TCM medical data. This heterogeneity is primarily characterized by multi-source and multi-modal data types, including image-based tongue diagnosis data, pulse diagnosis image data, and unstructured or semi-structured information from paper-based or electronic medical records. The effective aggregation, integration, and in-depth utilization of these diverse data forms remain a considerable challenge, especially in the digitization of vast historical paper records and images, which often still relies on inefficient manual extraction methods, thereby hindering the full potential of data-driven clinical research and application in TCM.

\begin{figure*}[htbp]
    \centering
    \includegraphics[width=0.9\textwidth]{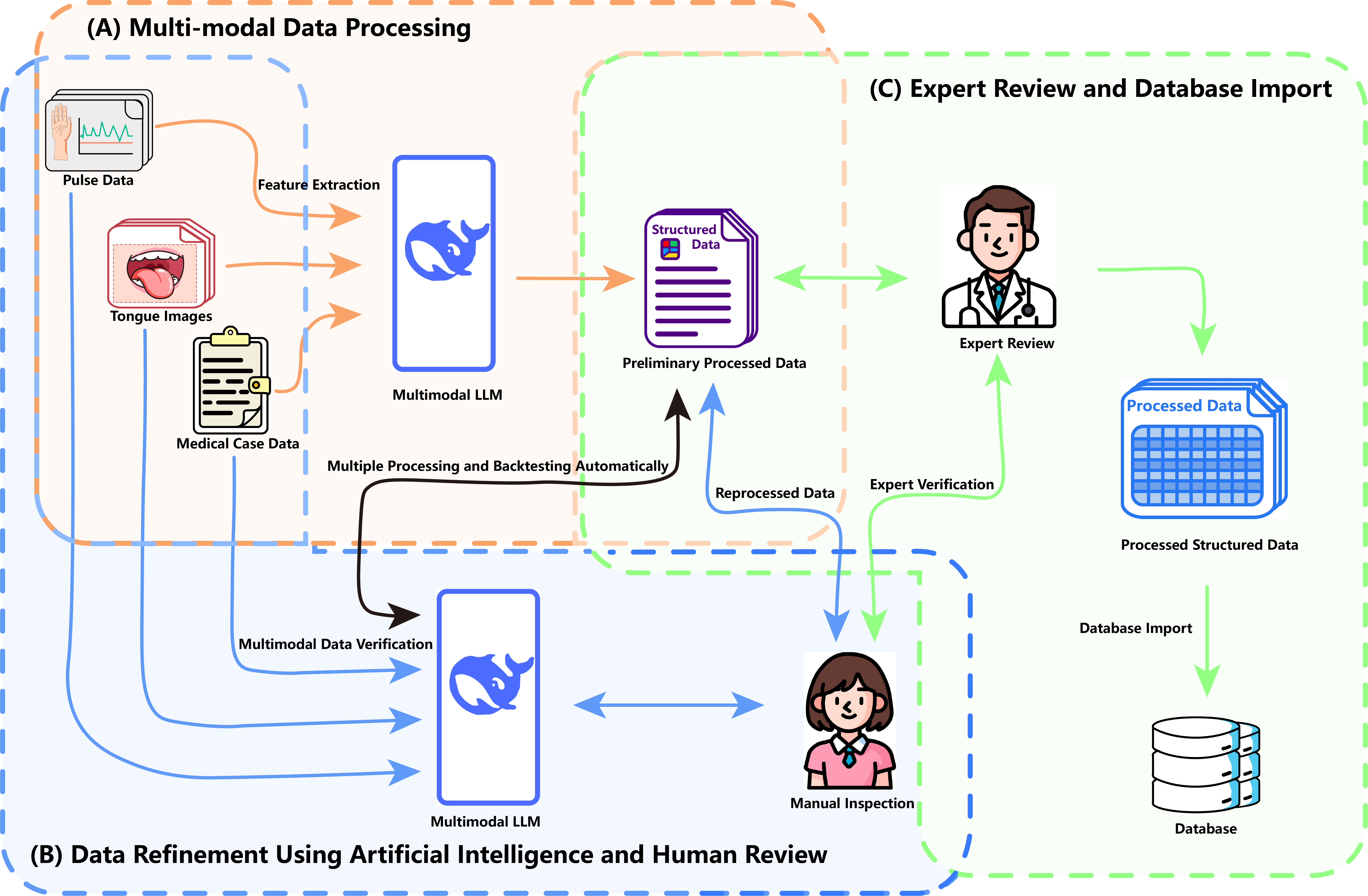}
    \caption{An overview of UFMD: (A) Extract features from multimodal data via MLLM templates. (B) Refine data through MLLM review and expert validation. (C) Store verified structured data in persistent database.}
    \label{fig:UFMD}
\end{figure*}

To address this problem, we propose a unified data fusion framework. The core of this framework involves the introduction of an advanced DeepSeek-OCR model~\cite{DeepSeek-OCR}. The contextual optical compression technology employed by this model enables the intelligent parsing of medical record images, image-based tongue diagnosis data, pulse diagnosis image data containing complex layouts and handwritten text, directly outputting high-quality structured data. Furthermore, we combine manual proofreading with expert verification in the framework to ensure the reliability and authenticity of the data. Fig.~\ref{fig:UFMD} illustrates the structure of UFMD. This approach achieves an "end-to-end" automated conversion of photos and paper-based medical record data, significantly enhancing the overall efficiency of extraction and integration from multimodal heterogeneous data to standardized, computable information, thereby laying a solid foundation for subsequent construction of high-quality TCM clinical datasets.

To mitigate data extraction anomalies and erroneous generation caused by hallucination in MLLM, we propose a novel semantic relevance verification mechanism. For an electronic medical record datum $D_{ex}$ extracted by MLLM and its corresponding raw text datum $D_{raw}$, we ensure the semantic relevance of the results by computing the cosine similarity between their sentence embeddings generated by a pre-trained Sentence-BERT (SBERT) model.

The Sentence-BERT model maps an input sentence to a fixed-dimensional dense vector space. Formally, the embedding of a sentence $S$ is given by:
\begin{equation}
\mathbf{e}_S = \text{SBERT}(S)
\end{equation}
where $\text{SBERT}(\cdot)$ denotes the forward pass of the Sentence-BERT model, and $\mathbf{e}_S$ is the resulting sentence embedding vector.

The semantic similarity between $D_{ex}$ and $D_{raw}$ is then calculated as the cosine similarity between their respective embeddings:
\begin{equation}
\begin{split}
\operatorname{sim}(D_{ex}, D_{raw})^{\text{SBERT}} 
    &= \cos(\mathbf{e}_{D_{ex}}, \mathbf{e}_{D_{raw}}) \\
    &= \frac{\mathbf{e}_{D_{ex}} \cdot \mathbf{e}_{D_{raw}}}{\|\mathbf{e}_{D_{ex}}\| \|\mathbf{e}_{D_{raw}}\|}
\end{split}
\end{equation}

\noindent The proposed verification mechanism operates as follows:
\begin{enumerate}
    \item \textbf{Text Preprocessing}: The extracted text $D_{ex}$ and the raw source text $D_{raw}$ undergo standard text preprocessing, including tokenization and cleaning (e.g., removing extra whitespaces, handling special characters).
    \item \textbf{Sentence Embedding Generation}: Each preprocessed text is fed into the Sentence-BERT model to obtain its fixed-dimensional semantic vector representation, $\mathbf{e}_{D_{ex}}$ and $\mathbf{e}_{D_{raw}}$.
    \item \textbf{Similarity Calculation}: The cosine similarity between the two embedding vectors is computed, yielding a score between -1 and 1, where scores closer to 1 indicate higher semantic relevance.
    \item \textbf{Relevance Thresholding}: A predefined threshold $\theta$ is applied to the similarity score. If $\operatorname{sim}(D_{ex}, D_{raw})^{\text{SBERT}} \geq \theta$, the extracted data $D_{ex}$ is considered semantically consistent with the source $D_{raw}$; otherwise, it is flagged for potential hallucination or error.
\end{enumerate}
This mechanism leverages the contextual understanding capabilities of Sentence-BERT, providing a robust assessment of relevance, especially for complex TCM narratives where paraphrasing and synonymy are common.

Furthermore, to ensure the accuracy and reliability of the core data content, we implemented a rigorous manual proofreading process. To further improve data processing efficiency, we introduced MLLM to constrain the uncertainty of generative data through information comparison, thereby enhancing data accuracy. This human-centric approach, augmented by AI assistance, involved domain experts systematically verifying the extracted and structured information against the original records. Through this meticulous manual proofreading, we ensured data accuracy while successfully completing the data transformation and structuring process for over 120,000 TCM clinical records.

\subsection{LingLan: SFT Model for TCM Diagnosis}

To enhance the diagnostic capability of a general-purpose LLM in TCM, we employ a Parameter-Efficient Fine-Tuning (PEFT) approach based on Low-Rank Adaptation (LoRA)~\cite{lora} in Qwen3-14B~\cite{qwen3}, which has high performance in Chinese tasks. This strategy allows us to effectively adapt the pre-trained model to the specialized knowledge and language patterns inherent in TCM clinical records, while avoiding the prohibitive computational cost associated with full parameter fine-tuning.

Our methodology emphasizes the systematic integration of multimodal diagnostic data, particularly focusing on tongue diagnosis description and palpation diagnosis description extracted from UFMD. Specifically, the training dataset was meticulously curated and partitioned into training and held-out test sets. Each clinical record was reformatted using a structured template that explicitly incorporates quantitative feature information from tongue images and pulse waveform images alongside inquiry information. This structured representation forces the model to establish clinically meaningful relationships between the objective findings from inspection and palpation and the subjective information from Inquiry, thereby directly aligning the learning objective with the holistic reasoning process of TCM diagnosis. By learning from this integrated representative descriptions, the model develops the capability to correlate specific tongue characteristics (e.g., tongue proper color, coating texture) and pulse patterns (e.g., wiry, thready) with corresponding TCM syndromes. This approach significantly improves the model's proficiency in analyzing patient symptoms and conducting syndrome differentiation based on the complete I-AOI-P diagnosis, moving beyond a reliance on textual descriptions of symptoms alone.

In our experimental setup, the fine-tuning process was efficiently configured using the PEFT library, with the key hyperparameters and hardware environment detailed in Table~\ref{tab:config_and_env} for a direct comparison against the BianCang-14B-Instruct model~\cite{biancang}. This comparative analysis highlights the superior efficiency of our PEFT-based approach, which significantly reduces training time and computational resource requirements while maintaining competitive performance metrics. The tabulated data clearly demonstrates the advantages of our method in terms of parameter efficiency and accelerated convergence.

\begin{table}[b]
\centering
\small
\caption{Configurations and Training Environment Comparison}
\label{tab:config_and_env}
\begin{tabular}{l|p{2cm}|p{2cm}}
\hline
\multicolumn{1}{c|}{\textbf{Parameter} /} & \textbf{LingLan-14B} & \textbf{BianCang-14B} \\
\multicolumn{1}{c|}{\textbf{Environment}} & \textbf{(Ours)} & \textbf{~\cite{biancang}}
\cr\hline\hline
Base Model & Qwen3-14B & Qwen2.5-14B \\
LoRA Rank $(r)$ & 8 & - \\
LoRA Alpha $(\alpha)$ & 32 & - \\
LoRA Dropout & 0.05 & - \\
Target Modules & \texttt{all-linear} & - \\
Learning Rate & $1 \times 10^{-4}$ & - \\
Batch Size & 1 & - \\
Training Epochs & 1 & 2 \\
Hardware (GPUs) & NVIDIA RTX 5090 32GB & 8 $\times$ NVIDIA A100 40GB \\
\hline
Training Time & 12 hours & 605 hours \\
\hline
\end{tabular}
\end{table}

\section{Experiment}
\subsection{Experimental settings}
\label{Es}
\subsubsection{Dataset}
We use a 20\% subset of data from the TCM medical record dataset as a reserved test set for our experiments. We employ stratified sampling to maintain consistency in class distribution, stratifying by diagnostic label to ensure that the proportion of each diagnosis is approximately the same in the test and training sets. We also handle rare classes: if a class has only one sample, it is excluded from the test set; if there are two samples, at least one is included. This prevents evaluation bias caused by a class appearing only in either the training or test set.

\begin{figure}[h] 
    \centering
    \includegraphics[width=0.8\linewidth]{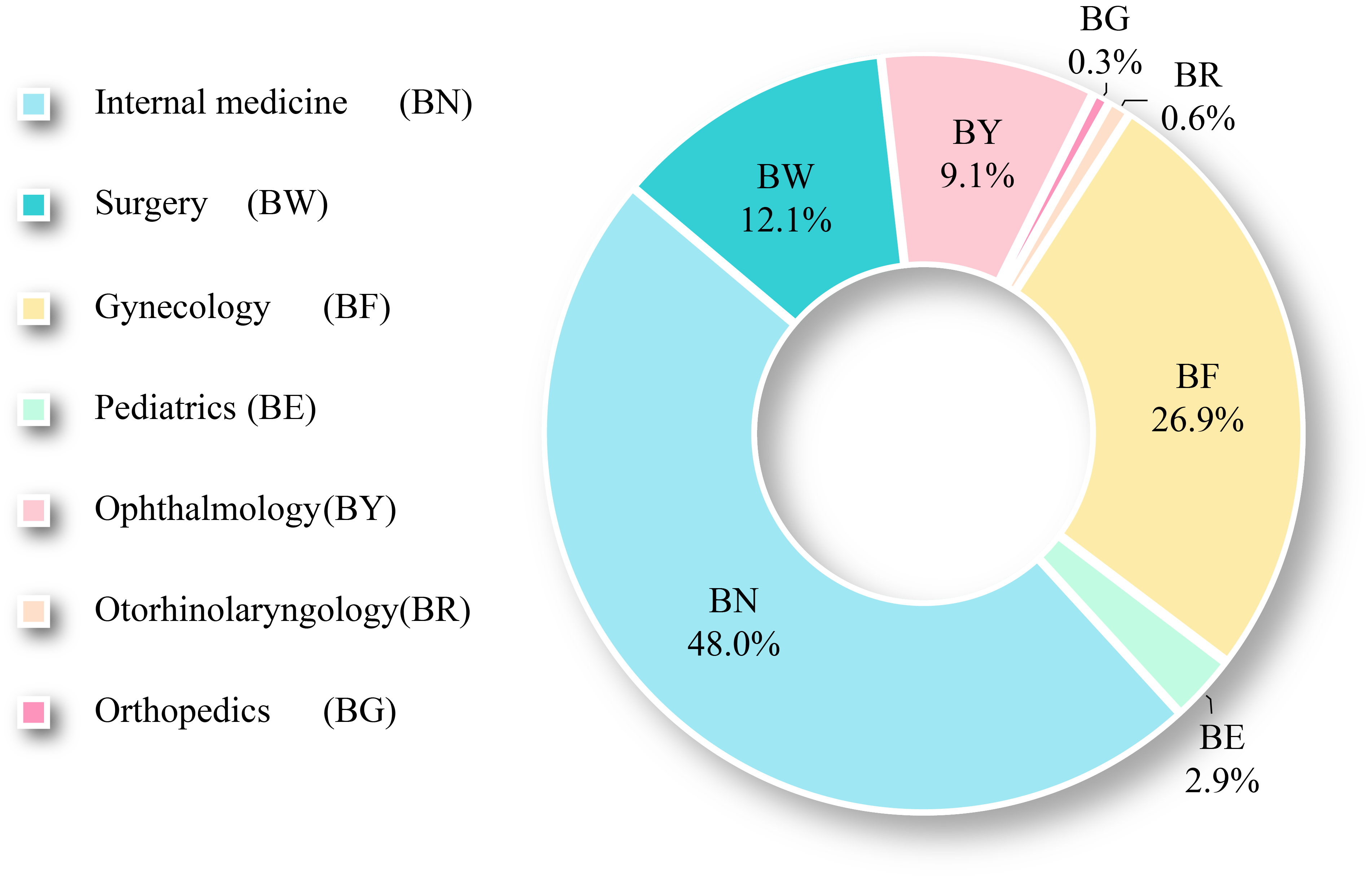}
    \caption{Distribution of Test Data}
    \label{fig:pie}
\end{figure}

To ensure the authority of the diagnostic classification, the diagnoses in this study are fully classified in strict accordance with\textit{the Classification and Codes of Diseases and Patterns of Traditional Chinese Medicine}(GB/T 15657-2021) approved by the China National Standardization Administration. The classification results and category statistics are shown in Fig.~\ref{fig:pie} and Table~\ref{tab:categories_of_test}.

% 标准的网址https://openstd.samr.gov.cn/bzgk/gb/newGbInfo?hcno=41FD9D06E5BE4F84EA1D8D07101BED2C

\begin{table}[b]
\centering
\caption{Classicifation Categories of Test Set}
\label{tab:categories_of_test}
\begin{tabular}{lll}
\hline
\textbf{Diagnosis type} & \textbf{Abbreviation} & \textbf{Count} \\
\hline
Internal medicine       &   BN   &   11708 \\
Surgery                 &   BW   &   699 \\
Gynecology              &   BF   &   6561 \\
Pediatrics              &   BE   &   154 \\
Ophthalmology           &   BY   &   84 \\
Otorhinolaryngology     &   BR   &   2208 \\
Orthopedics             &   BG   &   2958 \\
\hline
\end{tabular}
\end{table}

\subsubsection{Baselines}
We compare our fine-tuned model with the original model Qwen3-14B and open-source models with higher parameter levels. Among them, Qwen3-32B~\cite{qwen3}, DeepSeek-R1-Distill-32B and DeepSeek-R1(671B)~\cite{ds}, as general-purpose large-scale models trained on Chinese corpora, have better understanding of Chinese content and are therefore suitable choices for comparison. We also include a model specifically designed for TCM diagnosis, BianCang-14B-Instruct~\cite{biancang}, in the experiment to compare LingLan's performance with the professional TCM-LLM model.

\subsubsection{Metrics}  
In order to compare the performance, we choose to use basic metrics such as accuracy, recall, and F1-score to evaluate the model's fundamental TCM prediction ability. Furthermore, during our experiments, we found that general-purpose models, lacking quantifier constraints, tend to provide broader answers, leading to data inaccuracies. Therefore, based on the specific characteristics of TCM diagnosis, we designed the following two new metrics:
\begin{itemize}
\item \textbf{Unconstrained Diagnosis Indicator(UDI)}: This metric measures the predictive ability of LLM for specific patient examination results and individual needs in the absence of any constraints, aiming to test the predictive power of LLM in generalized scenarios.
\begin{equation}
\text{UDI} = \frac{U_{TP} + U_{TN}}{U_{TP} + U_{TN} + U_{FP} + U_{TN}}
\end{equation}

\item \textbf{Post Constrained Diagnosis Indicator(PCDI)}: To test the predictive ability of LLM for more specific symptoms, we added instruction constraints to the LLM during the testing process and observed its diagnostic predictive ability after providing these constraints.
\begin{equation}
\text{PCDI} = \frac{2P_{TP}}{P_{TP} + P_{TN} + P_{FP} + P_{TN}}
\end{equation}
\end{itemize}

\subsection{Results}
\subsubsection{Overall Performance}Table~\ref{tab:Exp_result} compares the performance of the fine-tuned model and the baseline models on the TCM medical record test set. The results clearly demonstrate that LingLan-14B model achieves state-of-the-art diagnosis performance, significantly outperforming all baseline models across all evaluation metrics, including Accuracy, Recall, F1-score, UDI, and PCDI. This substantial improvement, particularly the more than doubling in Accuracy and F1-score compared to some larger models, underscores the effectiveness of our fine-tuning approach for the specialized domain of TCM medical records. The superior performance establishes a strong foundation for the subsequent discussion on the specific contributions and implications of our method.

\begin{table}[b]
\caption{Experiment Result of Diagnosis Performance}
\setlength{\tabcolsep}{3pt}\renewcommand\arraystretch{1}
\centering
\begin{tabular}{l|lllll}
\hline
\label{tab:Exp_result}
Model                          & Acc.  & Recall & F1-score & UDI & PCDI \cr\hline
\hline
Qwen3-14B~\cite{qwen3}            & 0.3082 & 0.4109      & 0.4112        & 0.3106   & 0.2875    \\
Qwen3-32B~\cite{qwen3}            & 0.3324     & 0.4293      & 0.4098        & 0.3647   & 0.3312    \\
DeepSeek-R1-Distill-32B~\cite{ds}          & 0.2700     & 0.3774      & 0.3591        & 0.2824   & 0.2776    \\
DeepSeek-R1(671B)~\cite{ds}              & 0.3097     & 0.4356      & 0.4425        & 0.3627   & 0.3030    \\
BianCang-14B-Instruct~\cite{biancang}              & 0.3266 & 0.6951      & 0.6949        & 0.5015   & 0.4394    \\
\hline
\textbf{LingLan-14B (Ours)}       & \textbf{0.6272} &  \textbf{0.8203}      &  \textbf{0.8202}        &  \textbf{0.6325}   &  \textbf{0.6103}    \\
\hline
\end{tabular}
\end{table}

\begin{figure}[h]
    \centering
    % 第一个子图
    \begin{subfigure}{0.48\linewidth}
        \centering
        \includegraphics[width=\textwidth]{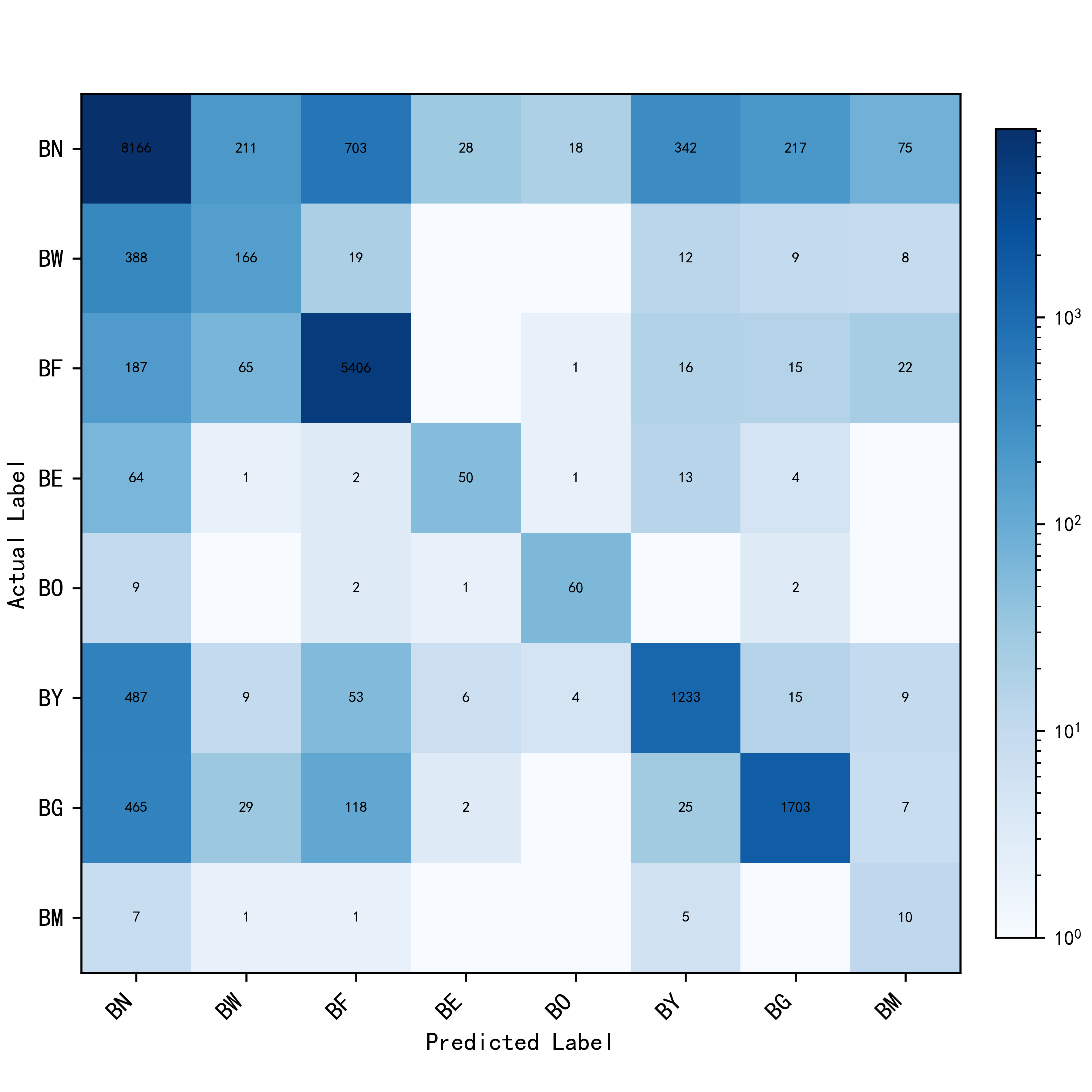}
        \caption{} % 子图标题留空，则自动显示为 (a)
        \label{fig:subfig_a} % 子图标签
    \end{subfigure}%
    \hfill
    % 第二个子图
    \begin{subfigure}{0.48\linewidth}
        \centering
        \includegraphics[width=\textwidth]{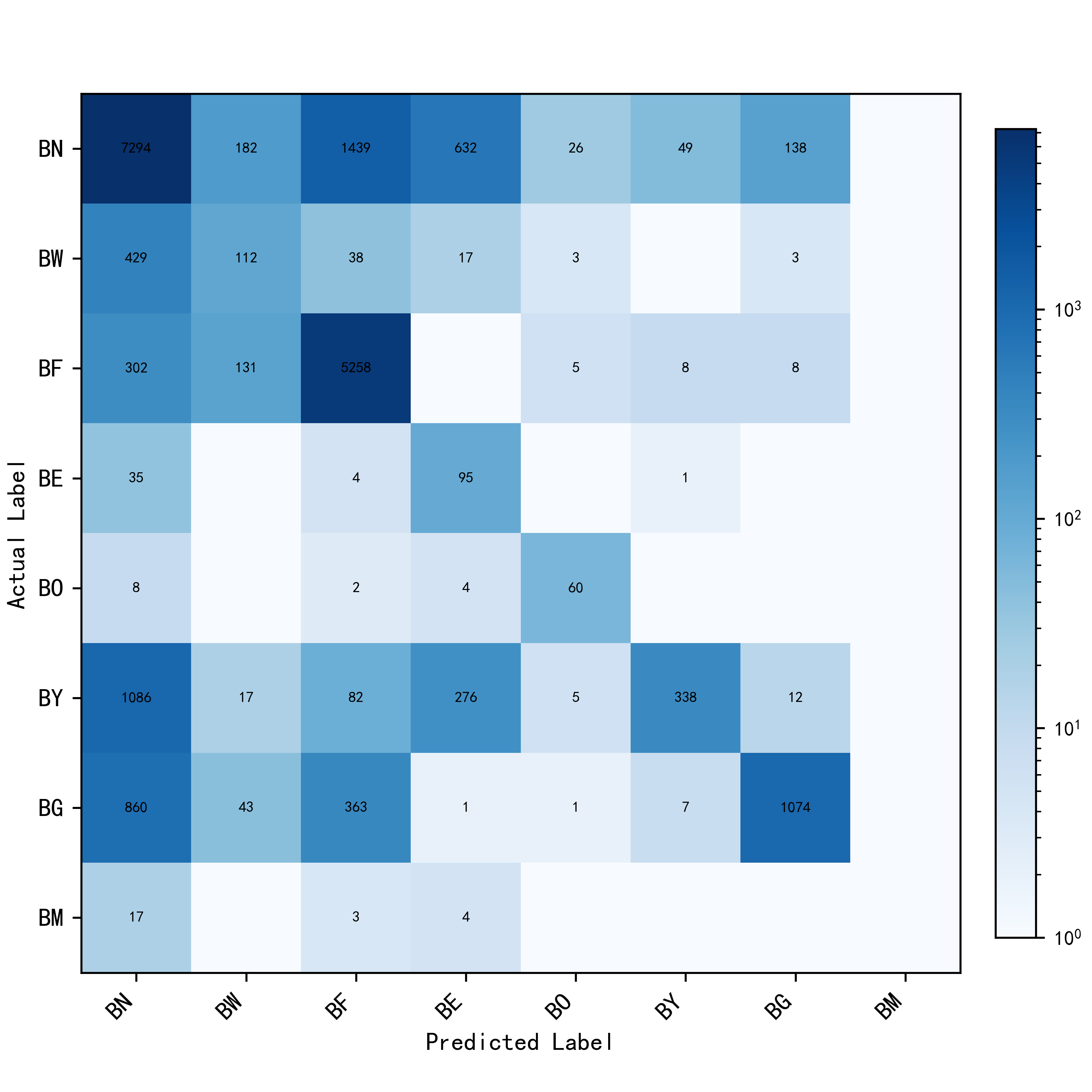}
        \caption{} % 自动显示为 (b)
        \label{fig:subfig_b}
    \end{subfigure}
    % 主标题
    \caption{Confusion matrix of (a) LingLan-14B (b) BianCang-14B-Instruct}
    \label{fig:confusion}
\end{figure}

To quantitatively evaluate the classification efficacy of our fine-tuned model, we present the confusion matrix heatmaps of our model and BianCang-14B-Instruct in Fig.~\ref{fig:confusion}. The matrices illustrate the alignment between the true labels (vertical axis) and the model's predicted labels (horizontal axis) across all categories (BM represents the misclassified label). A predominant concentration of high values along the main diagonal is immediately evident, signifying strong predictive accuracy. Statistics show that LingLan's recall accuracy reaches 82.03\%, compared to 69.51\% of BianCang-14B-Instruct. For more detail, our model correctly classified 2,850 instances for category BN, 5,406 for BF, and 1,703 for BG, as indicated by the intensely colored corresponding diagonal cells. While minor misclassifications are present in the off-diagonal elements—such as confusions between BN and BW (703 cases), BN and BY (487 cases), and BF and BG (1,103 cases)—their relatively low magnitudes compared to the diagonal values underscore the model's overall robustness. The clear diagonal dominance visualized by the color gradient confirms that our fine-tuned model achieves high classification performance, effectively learning discriminative features for accurate syndrome differentiation in TCM diagnostics.

\subsubsection{Ablation Study}We conduct an ablation study to validate the effectiveness of our novel training set structure in predicting TCM diagnosis. In addition, to better demonstrate the assistance of structured data in TCM diagnosis, we calculate the accuracy and recall of the diagnosis and syndrome differentiation results separately. We fine-tune the same baseline model Qwen3-14B with the same parameters using the raw data without any data structure processing, and test the model using the same test set. Table~\ref{tab:abla_Result} shows that the results of fine-tuning with structured data are significantly better than those without.

\begin{table}[b]
\caption{Ablation Experiment Result}
\setlength{\tabcolsep}{3.5pt}\renewcommand\arraystretch{1}
\centering
\begin{tabular}{l|llll}
\hline
\label{tab:abla_Result}
Model                          & Diag$_{Acc}$  & Diag$_{R}$ & Syn\_Diff$_{Acc}$ & Syn\_Diff$_{R}$  \cr\hline\hline
SFT$_{without}$           & 0.3790 & 0.6542      & 0.2460        & 0.4626       \\
SFT$_{with}$            & \textbf{0.6334}     & \textbf{0.8277}      & \textbf{0.6210}        & \textbf{0.6020}       \\
\hline
\end{tabular}
\end{table}

\section{Conclusion}
In this paper, we propose UFMD, an innovative MLLM-based framework for intelligent diagnosis in TCM, addressing the challenges of integrating holistic multimodal I-AOI-P diagnostic information with quantitative medical systems. We also present LingLan-14B, a TCM-specific LLM fine-tuned via supervised learning to emulate the I-AOI-P diagnostic workflow, which reaching SOTA diagnostic performance. Experimental results demonstrate that our method significantly enhances diagnostic accuracy, validating the effectiveness of our structured data integration and fine-tuning strategy.

\section{Future Work}
We must state that due to the involvement of a substantial amount of patient personal privacy data, the dataset utilized in this study cannot be made publicly available at this time. In the future, we endeavor to release an anonymized version of the data after undergoing a formal privacy security assessment and obtaining appropriate patient authorizations. Furthermore, owing to cybersecurity constraints, we were unable to empirically evaluate the performance of non-open-source models on TCM diagnosis task. Our future work will explore the feasibility of conducting security-compliant testing for such models without compromising patient data privacy. Additionally, while our fine-tuning methodology yielded significant improvements on smaller-scale models, we plan to extend its application to larger models in subsequent research to further validate its effectiveness and scalability.

% \section*{Acknowledgment}

\bibliographystyle{IEEEbib}
\bibliography{icme2026references}

\end{document}